\documentclass[]{spie}  %>>> use for US letter paper
\usepackage{amsmath,amsfonts,amssymb}
\usepackage{graphicx}
\usepackage[colorlinks=true, allcolors=blue]{hyperref}
\usepackage{braket}
\usepackage{breqn}
\usepackage{booktabs}
\usepackage{dcolumn}
\usepackage{enumitem,comment}
\usepackage[table]{xcolor}
\usepackage{blkarray}
\usepackage{bm}
\usepackage{float}

\usepackage{amsmath}
\usepackage{algorithm}
\usepackage{caption}
\usepackage{subcaption}
\allowdisplaybreaks

\usepackage{color}
\usepackage{graphicx}

\usepackage{amsmath,amsfonts,amssymb}
\usepackage{graphicx}

\definecolor{firstColor}{cmyk}{0.053333333,0.0088888888,0.0088888888,0.017777777}
\definecolor{secondColor}{cmyk}{0.01,0.01,0.11,0.02}
\def\rddots#1{\cdot^{\cdot^{\cdot^{#1}}}}

\title{Optimization problems with low SWaP tactical Computing }

\author[a]{Mee Seong Im, Ph.D.}
\author[b]{Venkat R. Dasari, Ph.D.} 
\author[c]{Lubjana Beshaj, Ph.D.}
\author[b]{Dale Shires}
\affil[a]{U.S. Military Academy, West Point, NY 10996}
\affil[b]{U.S. Army Research Laboratory, Aberdeen Proving Ground, MD 21005}
\affil[c]{Army Cyber Institute, West Point, NY 10996}

\authorinfo{Further author information: 
(Send correspondence to Venkat~R. Dasari) \\ 
Mee Seong Im: E-mail: meeseong.im@westpoint.edu \\
Venkat R. Dasari: E-mail: venkateswara.r.dasari.civ@mail.mil \\ 
Lubjana Beshaj: E-mail: lubjana.beshaj@westpoint.edu \\
Dale Shires: E-mail: dale.r.shires.civ@mail.mil
  } 
\begin{document} 
\maketitle

\begin{abstract}

In a resource-constrained, contested environment, computing resources need to be aware of possible size, weight, and power (SWaP) restrictions.  SWaP-aware computational efficiency depends upon optimization of computational resources and intelligent time versus efficiency tradeoffs in decision making.  In this paper we address the complexity of various optimization strategies related to low SWaP computing.  Due to these restrictions, only a small subset of less complicated and fast computable algorithms can be used for tactical, adaptive computing.

\end{abstract}


\section{Introduction}\label{sec:introduction}
The Army has put a significant amount of emphasis on size, weight, and power (SWaP) when it comes to deploying devices.  We optimize for SWaP in two ways.  First, at the hardware level we optimize by always staying at the leading edge of available technologies and rapidly employ them in the mission-critical extreme environments. Second, in the Army we rapidly adapt the latest processor technologies when they are available. We continue to leverage ongoing research for processors to be replaced with faster, smaller, and less power-consuming ones. Low SWaP computing platforms are widely deployed in mission-critical environments to perform a variety of computational tasks ranging from trivial to highly complex. Since processors have become more niche and specialized, they are now often heterogeneous.  In these heterogeneous computing environments with hyper mobility, achieving global optimization goals require special strategies and algorithms. Optimization of cost functions is very important in neural network-based computational frameworks like tensorflow and deep learning\cite{bagloee2018hybrid, deng2018learning, fischetti2018machine, joy2018flexible, xiao2017optimization}. 
 
\begin{figure}[htbp]
\begin{center}
\includegraphics[height=9cm, width=9.0cm]{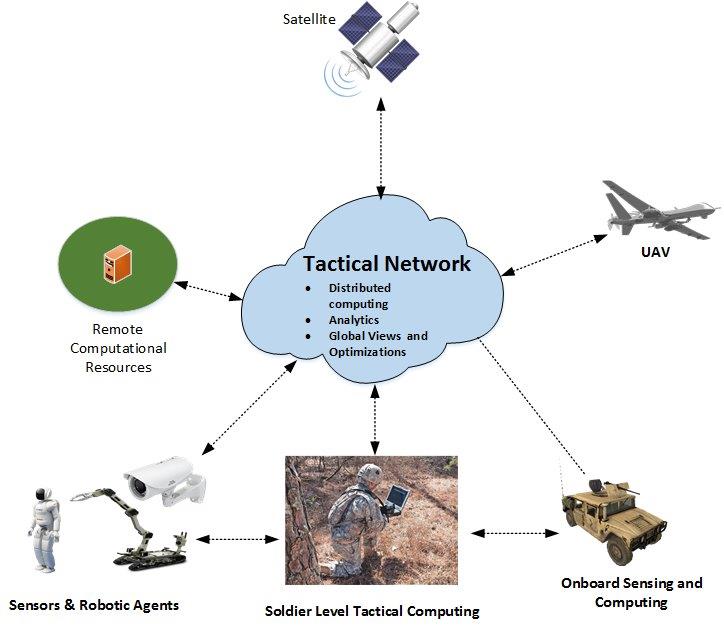}
\vspace{-0.25cm} 
\end{center}
\caption{The figure reflects an overview of low SWaP tactical computing platform deployment in mission-critical environments.}
\label{fig1}
\end{figure}

Tactical environments, common for Army operational concerns, greatly increase the need for computing platforms to be SWaP aware and optimized.  Though these platforms are SWaP constrained, with adaptive features and optimizations it is possible to deploy more functionality at a reasonable computational cost. This is possible through specialization based on the domain of the computing requirement.  Both industry and defense have been investing significant amount of resources in creating their electronic systems in sophisticated yet small-scale as needed to perform mission-critical functions using limited resources that are domain specific.  Another reason is to lessen the amount of weight of technological power, electronic displays, and communication sensors an infantryman must carry on a mission field. Along with novel adaptive computing strategies, efficient optimization algorithms will significantly increase the mission performance of the low SWaP platforms.

%%%%%%%%%%

\section{Computational complexity of adaptive mission platform}\label{section:computational-complexity-adaptive-mission}

We refer to literature by Aaronson~\cite{aaronson2013philosophers},  
Arora-Barak~\cite{arora2009computational},  
Gritzmann-Sturmfels~\cite{gritzmann1993minkowski},  and 
Hartmanis-Stearns~\cite{hartmanis1965computational} for an extensive background on computational complexity in mathematics and in theoretical computer science. 
Computational complexity is the study of the difficulty of solving problems about finite combinatorial objects, 
with deep ramifications in cryptography \cite{baker1975relativizations, marks2016universality}, 
algorithms\cite{cook2006p, shor2004progress, yan2002number}, 
game theory\cite{sahni1974computationally}, 
economics \cite{vives1984duopoly}, and  
artificial intelligence \cite{cook2006p, buss2012towards}. 
For example, given two integers $a$ and $b$, are they relatively prime, or given a list of cities and a distances between any two cities, what is the shortest route so that one visits each city and returns to the original position? 

Complexity theory aims to distinguish problems by measuring the level of difficulty of a feasibly decidable problem, i.e., it uses a conventional Turing machine whose computational steps are proportional to a polynomial function of the size of its input. 

Mathematicians and theoretical computer scientists has created time and space hierarchies, making complexity classes finer: 
\begin{equation}
P \subseteq NP \subseteq \text{PSpace} \subseteq \text{ExpTime} \subseteq \text{NExpTime}  
 \subseteq \text{ExpSpace} \subseteq 2\text{-}\text{ExpSpace} \subseteq \ldots \subseteq k\text{-}\text{ExpSpace},  
\end{equation}
where 
 $P$ is the deterministic polynomial time complexity class, 
 $NP$ is the class of languages decidable in polynomial time on a nondeterministic Turing machine, 
 $\text{PSpace}$ is the class of decision problems solvable by a Turing machine using a polynomial amount of space, 
 $\text{ExpTime}$ is the set of all decision problems that are solvable by a deterministic Turing machine in exponential running time $\mathcal{O}(2^{p(n)})$, where $p$ is a polynomial,   
 $\text{NExpTime}$ is the class of decision problems solvable by a nondeterministic Turing machine in exponential running time, 
 $\text{ExpSpace}$ is the class of decision problems solvable by a deterministic Turing machine in exponential running space $\mathcal{O}(2^{p(n)})$, where $p$ is a polynomial, 
 $2\text{-}\text{ExpSpace}$ is the class of decision problems solvable by a deterministic Turing machine in doubly-exponential running space $\mathcal{O}(2^{2^{p(n)}})$, where $p$ is a polynomial, and $j\text{-}\text{ExpSpace}$ is the class of decision problems solvable by a deterministic Turing machine in the order of the running space $\mathcal{O}\left(2^{\rddots{{p(n)}}}\right)$, where $2$ is raised to itself $j$-times. 
 We also have the following hierarchy of complexity classes: 
 \begin{equation}
P \subseteq NP \subseteq \text{PSpace} \subseteq \text{ExpTime} \\
 \subseteq \text{NExpTime} \subseteq \text{ExpSpace} \subseteq 2\text{-}\text{ExpTime},  
\end{equation}
where $2$-ExpTime is the set of all languages solvable by a deterministic Turing machine in the order of doubly-exponential $\mathcal{O}(2^{2^{p(n)}})$-time, where $p$ is a polynomial. In the latter hierarchy, the complexity classes are bounded by increasing levels of time bounds. 

It has been proven that 
 $P\subsetneq \text{ExpTime}$, $\text{PSpace}\subsetneq \text{ExpSpace}$, and 
 $NP \subsetneq \text{NExpTime}$ since exponential gaps contribute the difference when measuring the resource bound. 
 So if $P=NP$, then the class of problems solvable in exponential time are solvable by a nondeterministic Turing machine. 
 If $P\not= NP$, then there exist sparse computational problems in $NP$ that are not in $P$.

We refer to Dasari-Im-Geerhart~\cite{Dasari-Im-Geerhart-18} for an extensive discussion on mission computability in computational complexity.  
In short, the mission time complexity class is a subclass of $P$ consisting of problems that are solvable using an approximate algorithm in a finite sequence of steps, where a comparable result may be computed when given an instance from the set of inputs, thus providing mission effectiveness. 
The algorithm cost for the number of operations and the storage space are in the order of polynomial time $\mathcal{O}(n^k)$, where $k\geq 0$ (\cite[Defn.~1]{Dasari-Im-Geerhart-18}).

As an example, an algorithm may require $20$ minutes of computational time to arrive to its completion but we may only need $10$ minutes of the processing time as the human mind can deduce an answer to a problem after a certain amount of computational steps. Halting the algorithm before its completion may not give us an accurate answer but the speed of an algorithm by re-factorizing an old algorithm by inputting multiple mathematical equations, and then implement a computer programmable code via python, C++, and other languages will free-up our limited computational resources for other urgent needs.

Mission computable problems consistently return a feasible, approximated solution, which are characterized by their distance from their value to the optimal solution. 
A feasible solution is an approximate solution, which is classified by the value of its distance from the ideal (or optimal) solution. 
Therefore, the ratio of the approximated solution to the optimal solution is determined by the input size, the performance growth of the algorithm, and time limitations. Examples include linear and dynamic programming, localized search, randomized, and heuristic algorithms. 
Linear programming formulates the problem as a linear model, while dynamic programming constructs a solution from optimal solutions to sub-problems of the original language.
When given a potential solution,  
a local search algorithm seeks for a better neighboring solution, where the algorithm deforms an input until an improved and refined optimal solution is found, while randomized algorithm executes a random decision generator. 
Heuristic algorithms are encoded by exploratory, learning strategies, 
which offer no guarantee of a better, further reasonable solution.

\section{Optimization trade-offs}\label{section:trade-offs}

Computational complexity of optimization algorithms is often  non-linear and exponential  in nature  and that gets worse with the increase of the number of variables involved in the optimization process. In mission-critical environments, time to compute an acceptable answer is very critical. The conventional optimization methods might provide you with an accurate answer but the time to compute such an answer might take too long, reaching the $\mathcal{O}(n^{2 })$ complexity. Finding real global minimums and maximums are time consuming in non-linear optimization tasks.  Time vs. accuracy is a key trade-off when we are attempting to optimize computations in  SWaP challenged mission-critical environments. Often a fully optimized solution that takes a long time to compute is not necessary and a 90\% accurate answer that can be computed faster in mission time is  all that we need in order to solve a computational task\cite{sidiroglou2011managing}.  Optimization strategies using genetic algorithms will provide reasonable optimizations in shorter time by reducing the time complexity in optimization process\cite{feng1997using}.  This trade-off assumes more importance  for low SWaP platforms where conservation of computational resources is important in processing the computational tasks related to optimizations. Unnecessary computations  will consume resources and reduce the  overall computational capacity of the platforms.

With the advent of Artificial Intelligence (AI) and  Machine Learning (ML),  advanced intelligence is pushed to the edge of computing where the low SWaP platforms are predominantly deployed.  Trained neural network  models are pushed to the edge  for image recognition and real-time video analytics. Most of AI \& ML in the edge of  computing is related inference  because of its low computational demands. However, new strategies to train the models in the computing edge  by distributing the task among a cluster of edge nodes is also considered to reduce the time complexity in training a model using a large set of data. Intelligent optimization strategies coupled with AI will be of high use in low SWaP tactical computing environments\cite{sundararaj2005optimization}.

In the following two sections, we further explore optimization trade-offs by investigating resource constraints on computational efficiency in a heterogeneous computational platform and different combination of optimizing configurations and sensitivity to incremental modifications. 
 
\subsection{Resource constraints on computational efficiency in a heterogeneous computational platform} 

An adaptable computational platform allocates algorithms and technology congestion protocols using behavior models and network efficiency of fairness characterization.22, 23 The algorithms, in fact, make an optimal linear or nonlinear decision by considering objective functions together with constraint inequalities to broaden the scope
for a successful military operation.
Computational constraints, such as energy and time interference, influence the performance of a multi- computational platform in contested and congested environments, and are compelling the need for a constraints- aware adaptive computation framework.

Dasari-Im-Geerhart~\cite[Section 4]{Dasari-Im-Geerhart-18} discusses mission effectiveness by varying the resources to each application as the needs change. 
That is, if too many computational jobs are assigned to an array of cores, an algorithm is inherently designed to allow for failure. So the need for the algorithm to be programmed to minimize the number of failed jobs is highly desirable, and such design is called constraints-aware distributed computing. If one considers the computational complexity in relation to mission requirements, even the polynomial algorithms fail to compute within mission time without additional optimizations and adjustments to the algorithm itself in some cases. 

For example, applying image analysis to functioning security cameras may consume more local resources than the amount that is available. So the application dedicated to analyzing the images would give a time-to-complete restriction on each frame, but most frames may fail the time restrictions and the cores may terminate most jobs immediately. In such cases computational off-loading and distributed computing strategies should be considered for completing those computationally intense image and video analytics.
Leading edge adaptive computing platforms are equipped with adaptive computing functions to optimize computations in low SWaP conditions. They are also capable of intelligently decomposing complex computational tasks into smaller tasks and distributing them to different nodes to cooperatively compute the task in mission time.
Locally incomputable jobs are handed to an AI-enabled distributed computing decision-making engine that can distribute the task to a cluster of connected nodes. A global optimizer will guide the decision-making engine to determine which jobs should be scheduled or rescheduled. Priority labels can be attached to these distributed jobs to convey the urgency of the computation to the remote nodes.

A difference between the local optimizer and the global optimizer is that computational size must be varied to fit the computational resources for the local optimizer. And the next step is to vary the allocation of resources to competing applications. As a continuation of the above example, image analysis across multiple security cameras should be considered a single image analysis application while another application may be attempting to apply machine learning to detect hostile agents from the data provided by the security cameras, thus maximizing mission effectiveness by varying the resources to each application as the needs change. 
In this example, priority access should be given to image analysis in hostile zones while machine learning should be given flexible timing constraints to allow for the jobs to be scheduled on HPC machines located distant from hostile zones. 

\subsection{Different combination of optimizing configurations and sensitivity to incremental modifications}\label{subsection:combination-configurations}

There are many possible combinations of optimizations and configurations which cause optimization trade-offs. For example, a warp, which is also known as wavefront, is the smallest executable unit of code, i.e., it processes a single instruction over all of the threads in it at the same time. A programming abstraction known as a thread block represents a group of threads that can be executed in series or in parallel. A different combination of optimized configurations exists when increasing from a few large thread blocks or from numerous small thread blocks.
The impact on the different types of platform includes that larger thread blocks have better data locality where the actual data resides on the node, whereas the many small thread blocks move large amounts of computable data to computation. This minimizes network congestion, increasing the overall utility of the system. Increasing
the larger thread blocks also has higher thread synchronization overhead where a task spends idle time waiting for the completion of another task. This wastes the core on which the task is executing since time greater than a few microseconds spent in wait functions represents synchronization overhead. Finally, larger thread blocks potentially waste more thread space on each multiprocessor due to underutilization.

In the case of sensitivity to incremental modifications,  it has been shown by Makowski and Wierzbicki that post-optimization problem analysis has limited applications to actual problems that are naturally complex in nature \cite{makowski2003modeling}. 

\section{Hybrid approach}\label{section:hybrid-approach}

 In order to reduce the time complexity in optimization, hybrid optimization techniques are used  and a combination of  AI, genetic and  integer linear programming (ILP) models are used  \cite{valouxis2000hybrid,wang2001effective}.

 In the previous sections we addressed the complexity of various optimization strategies related to SWaP computing and hence only a small subset of less complicated and fast computable algorithms can be used for tactical adaptive computing. In this paper we develop a hybrid method based on machine-learning and optimization that makes SWaP technology in mission-critical tactical environments much more efficient. 

 Machine learning  and mathematical optimization have been combined in different ways in the past.  One of the popular applications is the so-called Perspective Analytic field which uses machine learning to predict a phenomenon in the future and then mathematical optimization to optimize an objective over that prediction, see Bertsimas-Kallus~\cite{berstsimaskallus} for more details. Fischetti-Fracaro~\cite{fischetti2018machine} investigate if machine learning trained on a large number of optimized solutions, could accurately estimate the value of  the optimized solution  for new instances. Their model was trained on real world data of offshore wind parks and their results show that machine learning is able to efficiently estimate the value of optimized instances for the offshore wind farm layout problem.  

Bagloee-Asadi-Sarvi-Patriksson use a hybrid machine learning and optimization method is used to solve bi-level problems~\cite{bagloee2018hybrid}. The hybrid method transforms the original problem to an integer linear programing problem based on a supervised learning technique and a tractable nonlinear problem.

We produce an optimization technique for constraints-aware computational platform in order to minimize abandoned jobs on clusters of local machines, and to prioritize jobs based on their imperative level. That is, with tactical computing in mind, 
we combine ILP, genetic and AI algorithms to create a hybrid algorithm to derive an optimized mathematical function to describe a hybrid optimization applied to a SWaP computational model.

We consider a computational cluster with a job scheduling system for batch jobs and remote desktops for graphical applications, i.e., a constraints-aware computational platform, and discuss a way to optimize the heterogeneous computing environment. 
A tactical network consists of distributed, computing, analytics, global views, and optimizations, all of which need to be balanced within the computing power.  Let the variables $x_1,x_2,\ldots, x_n$ denote technological tools using a SWaP tactical platform, all of which require computational power. 

Consider the following formulated model 
\begin{equation}
\text{Usage}(t) = P(t) -  \sum_{i=1}^{n} M_i(t_i) - M'_{[n]}(t_1,\ldots, t_n), 
\end{equation} 
where $P(t)$ denotes the total amount of available processing power after time $t$,  
$M_i(t_i)$ is the amount of computational power required for mission $x_i$ after time $t_i$, 
$M'_{[n]}(t_1,\ldots, t_n)$ is the inter-related required processing power for $x_1,\ldots, x_n$, where the usage of multiple functionalities at a given time affects the amount of available computational power (here, $[n]:=\{1,2,\ldots, n\}$), 
and 
$\text{Usage}(t)$ denotes the net amount of power available at a given time. We note that $M_i$ and $M'_{[n]}$ need not be linear or continuous functions although they may be approximated as linear programmable functions over the duration of an infinitesimal time. Furthermore, the energy needed for mission $x_i$ can begin to use up the processing power at a later time than the energy required for mission $x_j$, where $i\not=j$, resulting in the need for $n$ time variables.

If the initial time constraints are
\begin{equation}
\begin{split}
M_i(t_i) &\leq C_i  \:\: \mbox{ and }\:\:  t_i \geq 0 \quad \mbox{ where }1\leq i\leq n, \\ 
\end{split}
\end{equation}
then we modify the constraints as 
\begin{equation}\label{eqn:modified-constraints}
\begin{split}
M_i(t_i) &\leq p_i  C_i \:\: \mbox{ and }\:\:  t_i \geq 0   \quad \mbox{ where }1\leq i\leq n,  \\ 
\end{split}
\end{equation}
and where $0< p_i<1$ since by the definition of mission computable polynomial time complexity class\cite[Defn.~1]{Dasari-Im-Geerhart-18}, only $p_i(100)\%$ of the computation $M_i(t_i)$ for $x_i$ needs to be completed in order to complete a successful mission. 
Note that we may rescale the values of $t$ and $t_i$ to take values in the set of integers, resulting in a discretized ILP model. 

Now, since some of the computations $x_1,\ldots, x_n$ take priority over the others (these are the effects of the genetic importance of some of the computations), we need to input their priority into the intelligent constraints-aware platform. So suppose we rate them in the following way: 
\begin{equation}\label{eqn:rating}
R_{1,[n]}(x_1,\ldots, x_n;t_1,\ldots, t_n) \geq  R_{2,[n]}(x_1,\ldots, x_n; t_1,\ldots, t_n)    
\geq \ldots \geq  R_{n,[n]}(x_1,\ldots, x_n; t_1,\ldots, t_n),    
\end{equation}
where a numerical value assigned to $R_{j,[n]}$ depends on the preferential importance of $x_1,\ldots, x_n$ and the amount of completed computation $t_1,\ldots, t_n$. 
Let $N=R_{1,[n]}(x_1,\ldots, x_n;t_1,\ldots, t_n) $, the maximum value. 
We note that the genetic importance may take into account the affected region and the individual users. Thus the ordering of the priority of $x_1,\ldots, x_n$ may be moved around and be SWaPped within the executable time, in particular, when some of the computations can be approximated and thus no longer need to use up the limited resources (see \cite[Defn.~1]{Dasari-Im-Geerhart-18}), or when technologies $x_i$ are removed or more are added to the SWaP platform. 

Let $R_{j_i,[n]}$ be the rating corresponding to $x_i$ at time $t_i$. 
Implementing the rating~\eqref{eqn:rating} gives us the general model 
\begin{equation}
\text{Usage}(t) = P(t) -  \sum_{i=1}^{n} 
	\frac{R_{j_i,[n]}(x_1,\ldots, x_n;t_1,\ldots, t_n) }{N}M_i(t_i)  
		- M'_{[n]}(t_1,\ldots, t_n),   
\end{equation}
 subject to the constraints 
 \begin{equation}\label{eqn:modified-constraints-fin-model}
\begin{split}
M_i(t_i) &\leq p_i  C_i \:\: \mbox{ and }\:\:  t_i \geq 0   \quad \mbox{ where }1\leq i\leq n.  \\ 
\end{split}
\end{equation}
In our definition of mission complexity~\cite{Dasari-Im-Geerhart-18}, we restrict to computations that are solvable in polynomial time. This imples that for each $1\leq i\leq n$,  $M_i(t_i)$ is computable in the order of $\mathcal{O}(n^k)$, where $k$ is some integer.

 \section{Conclusion}
The mission computational complexity class consists of mission-ready problems that are computable in mission time. Mission times are determined by the context in which the computation is used and the completion time of the task to determine the importance and the usefulness of the computation. Computational complexity of polynomial problems does not directly indicate the mission computability of such problems. Polynomial algorithms that cannot be computed in mission time will require additional conditions and optimization until they satisfy mission computability requirements. An intelligent constraints-aware adaptive computing framework can change computational behaviors of adaptive computing platforms in response to available resources and the complexity of the computational tasks. Taking all the mission requirements into consideration while making computing decisions is important in order to provide a mission deployability state of any polynomial algorithms. The optimization model described in this paper is well-suited for implementing in resource-constrained tactical environments and can easily be coupled to adaptive computing engines to improve the computational efficiency of the tactical computing platforms.

We have formulated an optimization model that merges mission computability18 and a low SWaP tactical platform using n general sources that use the computational power. The model takes into account multiple computations with independent running time, the possible loss of processing power due to an overload demand,
computability and optimization within mission time, and prioritized jobs using a genetic rating.
Our future work includes an application of our model to a local SWaP platform to create a system of equations, applicable to the localized mission requirements.

\acknowledgments % equivalent to \section*{ACKNOWLEDGMENTS}       
 
This work is supported by a research collaboration between the U.S. Army Research Laboratory and U.S. Military Academy. 
M.S.I. is supported by National Research Council in Washington D.C.

% References
\bibliography{optimization} % bibliography data in report.bib
\bibliographystyle{spiebib} % makes bibtex use spiebib.bst

\end{document}